\documentclass[sigconf, nonacm]{acmart}
\usepackage[utf8]{inputenc}
\usepackage{subcaption}
\AtBeginDocument{%
  }

\setcopyright{acmlicensed}
\copyrightyear{2026}
\acmYear{2026}
\acmDOI{XXXXXXX.XXXXXXX}
\acmConference[Conference acronym 'XX]{Make sure to enter the correct
  conference title from your rights confirmation email}{June 03--05,
  2018}{Woodstock, NY}

\acmISBN{978-1-4503-XXXX-X/2018/06}

\begin{document}

\title[A Study on the Effects of Finger Movements on a Humanoid Hugging Robot]{A Study on the Effects of Finger Movements \\on a Humanoid Hugging Robot}

\author{Daigo Suhara}
\affiliation{
  \institution{Kansai University}
  \city{Takatsuki}
  \state{Osaka}
  \country{Japan}}
\email{k368213@kansai-u.ac.jp}

\author{Tomoko Yonezawa}
\affiliation{
  \institution{Kansai University}
  \city{Takatsuki}
  \state{Osaka}
  \country{Japan}}
\email{yone@kansai-u.ac.jp}

\renewcommand{\shortauthors}{Suhara et al.}

\begin{abstract}
This study aimed to examine the impact of finger movements in hugging robots on users by experimentally investigating how the presence and patterns of such movements affect users psychologically. The results suggest that the presence and patterns of finger movements influence the sensations users seek from the robot; specifically, repetitive movements and asynchronous movements across multiple fingers tended to enhance these sensations. Additionally, simultaneous movement of all fingers and gentle hugging tended to evoke a greater sense of care. Conversely, no differences were observed regarding resistance to physical contact initiated by the robot.
\end{abstract}

%%
%% The code below is generated by the tool at http://dl.acm.org/ccs.cfm.
%% Please copy and paste the code instead of the example below.
%%

%%\ccsdesc[500]{Human-centered computing~Human computer interaction (HCI)}
%%\ccsdesc[300]{Computer systems organization~Robotics}

%% A "teaser" image appears between the author and affiliation
%% information and the body of the document, and typically spans the
%% page.

%\received{20 February 2007}
%\received[revised]{12 March 2009}
%\received[accepted]{5 June 2009}

%%
%% This command processes the author and affiliation and title
%% information and builds the first part of the formatted document.
\maketitle

\section{Introduction}
In recent years, alongside advancements in human-robot interaction, the roles of robots have expanded beyond the industrial sector into fields involving close human contact, such as healthcare, welfare, and education. Consequently, there is a growing demand for designs that consider human psychology and emotions, in addition to safety and reliability.

The act of hugging, in particular, has been shown to contribute to a sense of security and stress reduction \cite{field, jakubiak}. Human-to-human hugging is known to stimulate oxytocin secretion and foster feelings of trust and intimacy \cite{kosfeld}; furthermore, subtle physical touch—known as the "Midas touch" effect—has been demonstrated to positively influence human behavior and affective states \cite{midas}. Expectations are high for applying these tactile and relational effects to human-robot interactions \cite{hoffmann}.

While research on hugging robots and robotic tactile interactions has been conducted previously, most studies have focused on the movement of the arms, torso, or overall physical softness and size, paying insufficient attention to finger movements \cite{shiomi, six-hug, probo, huggable}. However, in human interactions, subtle finger movements—such as stroking or grasping—are believed to significantly influence psychological impact and relational perception \cite{onishi}. 

Therefore, this study aims to clarify how variations in a hugging robot's finger movements affect the user; to this end, we developed a robot capable of independent finger movement and conducted an evaluation experiment.

\section{Proposed System}
\begin{figure}[htbp]
    \centering
    \includegraphics[width=0.45\columnwidth]{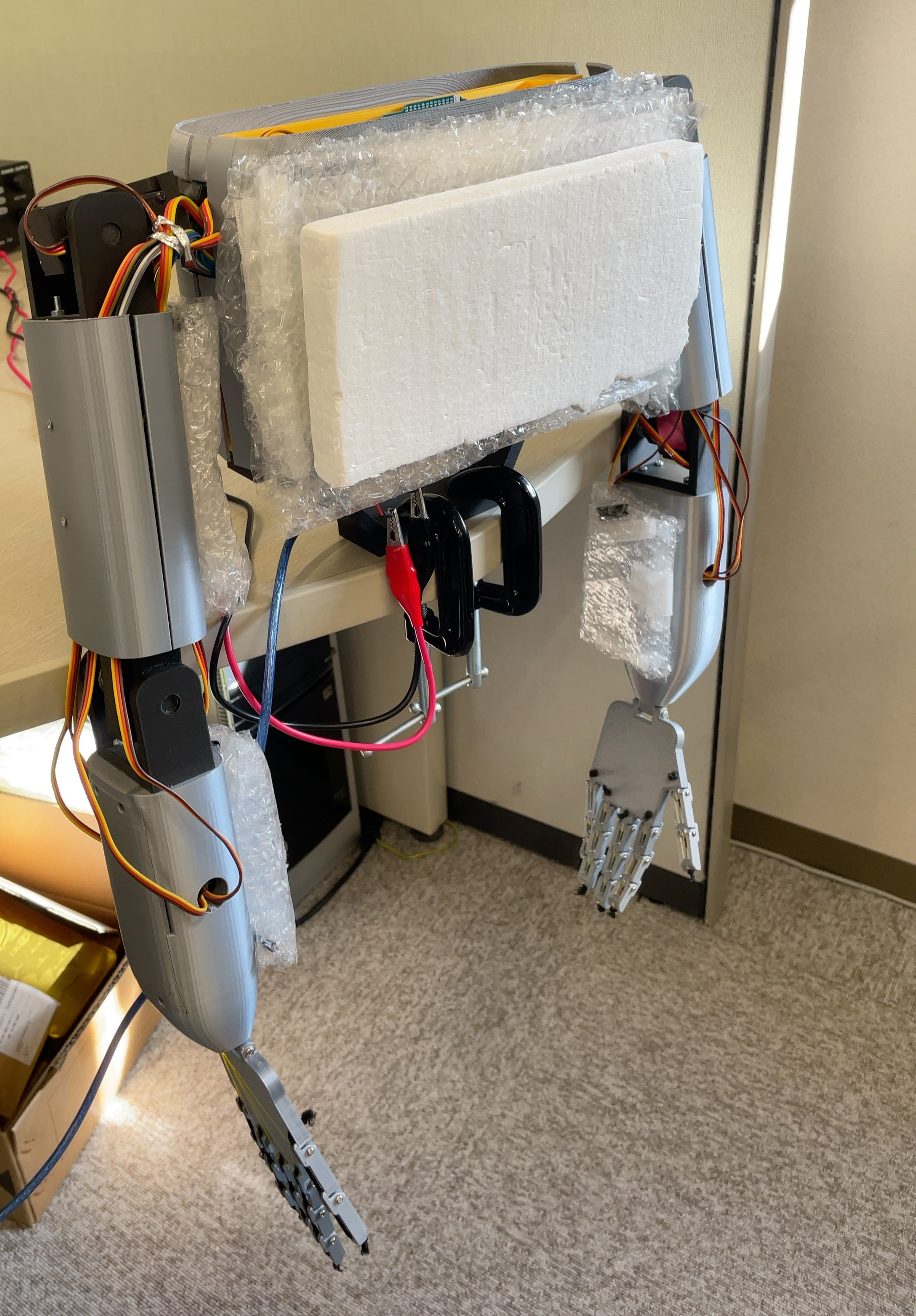}
    \caption{Hugging robot}
    \label{hugging-robot}
\end{figure}

\begin{figure}[htbp]
    \centering
    \includegraphics[width=0.3\columnwidth, angle=-90]{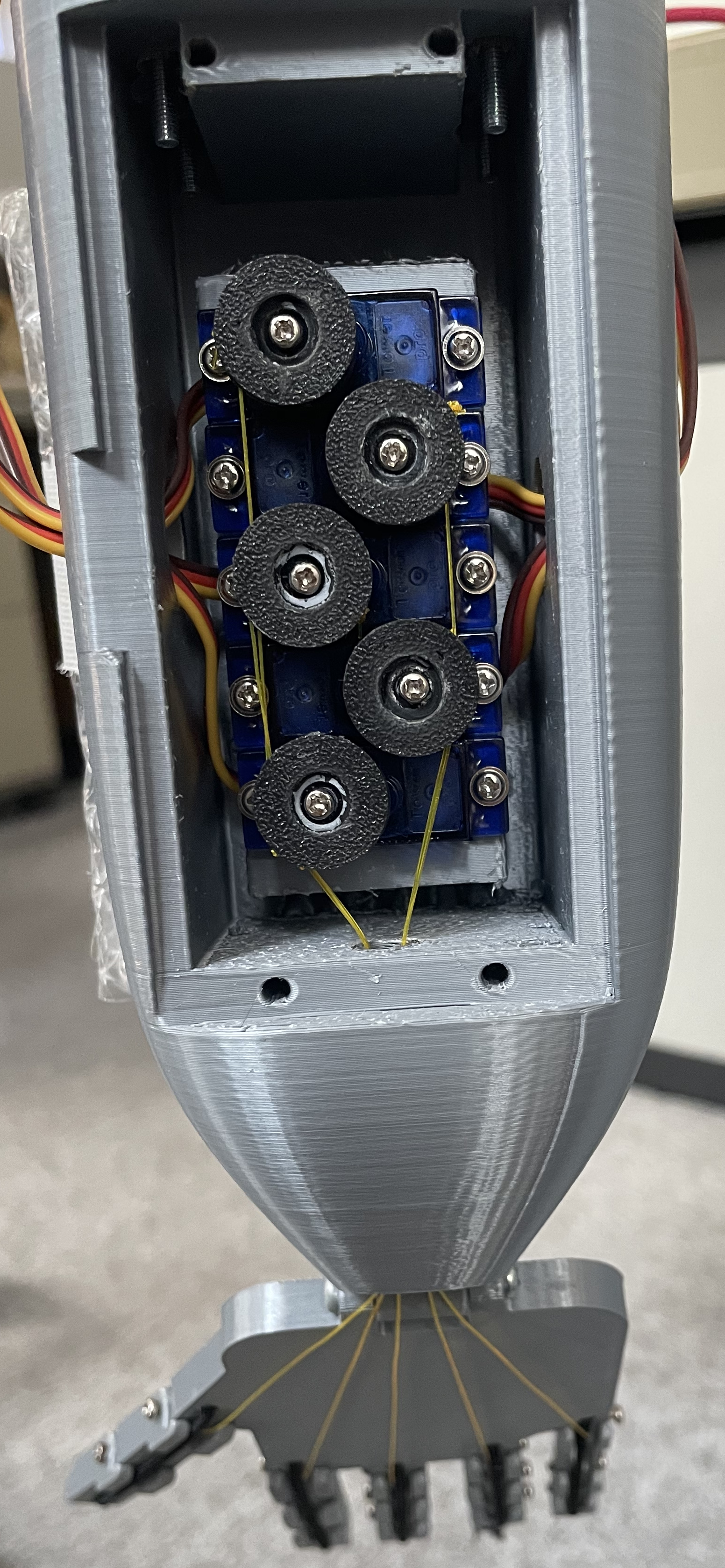}
    \caption{Robot arm}
    \label{robot-arm}
\end{figure}

Figure \ref{hugging-robot} shows the external appearance of the hugging robot, and Figure \ref{robot-arm} illustrates the structure of the robot arm. In this study, we developed a semi-humanoid hugging robot equipped with a three-degree-of-freedom arm and a robotic hand featuring five independently driven fingers. To avoid the "uncanny valley" effect, the robot is designed without a face\cite{mori}. A 3D printer was used to fabricate the frame and exterior components, utilizing PLA Silk material. Conversely, carbon fiber-reinforced PLA was selected for the servo motor brackets—which are subject to high loads—to balance cost-efficiency and durability.

\begin{figure}[htbp]
  \centering
  
  % 1つめのサブ図
  \begin{subfigure}{\linewidth}
    \centering
    \includegraphics[width=0.5\linewidth]{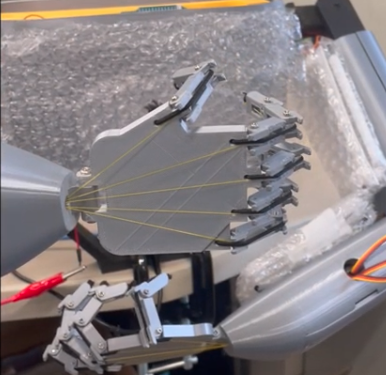}
    \caption{Simultaneous} % subfigure環境の中では \caption を使います
    \label{simultaneous}
  \end{subfigure}
  
  \vspace{5mm}
  
  % 2つめのサブ図
  \begin{subfigure}{\linewidth}
    \centering
    \includegraphics[width=0.5\linewidth]{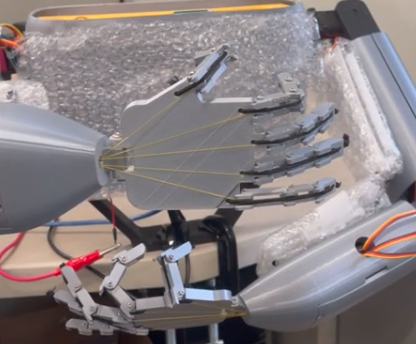}
    \caption{Independent}
    \label{independent}
  \end{subfigure}

  \caption{Independent control of robot fingers}
  \label{finger-movements}
\end{figure}

The joint structure and opening/closing movements of the robotic hand mimic human anatomy. The hand supports both simultaneous movement of all fingers (Fig. \ref{simultaneous}) and independent movement of each finger (Fig. \ref{independent}); the latter enables the testing of hypotheses regarding complex communication through independent finger articulation, allowing for the selection of specific movements based on context and intent.

These hugging motions were created using Bottango and are executed via a Bottango bootloader installed on an Arduino Mega.

\section{Experiment}

\subsection{Participants and Procedures}
\textbf{Participants: }
20 people aged 20 to 27 (15 men, 5 women; mean, standard deviation = 22.05, 1.356) participated in this experiment.
\textbf{Procedures: }
Participants sat facing the developed robot, which was positioned such that its shoulders aligned with the height of the user's underarms, allowing the user to look down at the robot while being embraced. Consequently, the evaluation was conducted face-to-face without any preceding or subsequent scenarios; participants used Google Forms to rate their impressions of the robot's frontal embrace on a 7-point Likert scale for each question item.

\subsection{Experiment 1: presence or absence of finger movement}

\textbf{Overview: }
In Experiment 1, in order to verify the psychological influence of the presence of finger movements, we conducted a within-subjects experiment with one factor and two conditions: Factor A: finger movements (A1: present, A2: absent). The question items are as follows.

\begin{itemize}
\item [Q1]: I felt thoughtfulness from the robot
\item [Q2]: I felt a strong demand coming from the robot
\item [Q3]: I felt uncomfortable about being touched by a robot
\end{itemize}

\begin{figure}[htbp]
    \centering
    \includegraphics[width=0.9\linewidth]{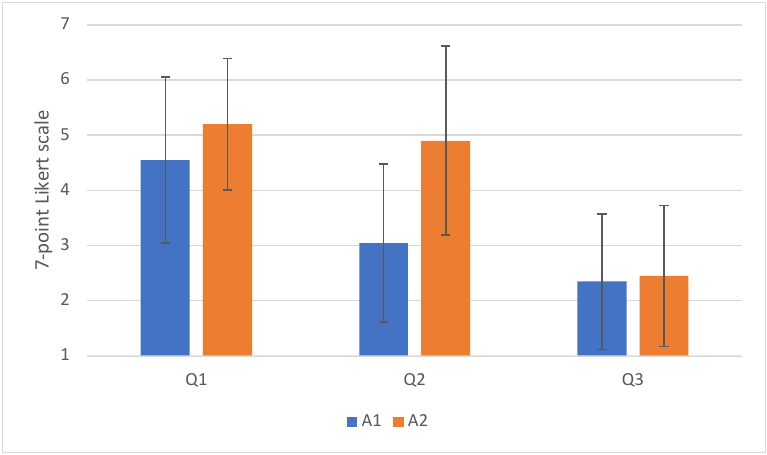}
    \caption{Mean and Standard Deviation (Experiment 1)}
    \label{test1-graph}
\end{figure}

\noindent\textbf{Results: }
A t-test was performed on the response data obtained for each question item at a significance level of $a$ = 0.05. The test results are presented in Table \ref{table:t-test} and Figure \ref{test1-graph}.

\begin{table}[htbp]
  \centering
  \caption{Results of the t-test}
  \label{table:t-test}
  \footnotesize
  \setlength{\tabcolsep}{8pt}
  \begin{tabular}{l cc}
    \toprule
    & $t$ & $p$ \\
    \midrule
    Q1 & -1.898 & .073 \\
    Q2 & -5.176 & $< .001$* \\
    Q3 & -0.400 & .694 \\
    \bottomrule
    \multicolumn{3}{l}{\scriptsize *$p < .001$}
  \end{tabular}
\end{table}

\subsection{Experiment 2: differences in finger and hand movements}

\textbf{Overview: }In Experiment 2, a within-subjects experiment involving three factors and eight conditions was conducted to examine the psychological effects of variations in finger and hand movement patterns during an embrace and combinations of embrace intensity. The factors were: Factor A: simultaneity of finger movements(A1: simultaneous, A2: asynchronous), Factor B: number of finger movements(B1: once, B2: multiple times), and Factor C: embrace intensity(C1: weak, C2: strong).The questionnaire items are as follows:
\begin{itemize}
\item [Q1]: I perceived a childlike quality in the robot
\item [Q2]: I felt a sense of security with the robot
\item [Q3]: I felt reluctant to be touched by the robot
\item [Q4]: I felt as though the robot was strongly seeking me out
\item [Q5]: I sensed thoughtfulness from the robot
\end{itemize}

\begin{figure*}[h]
  \begin{center}
    \includegraphics[width=160mm]{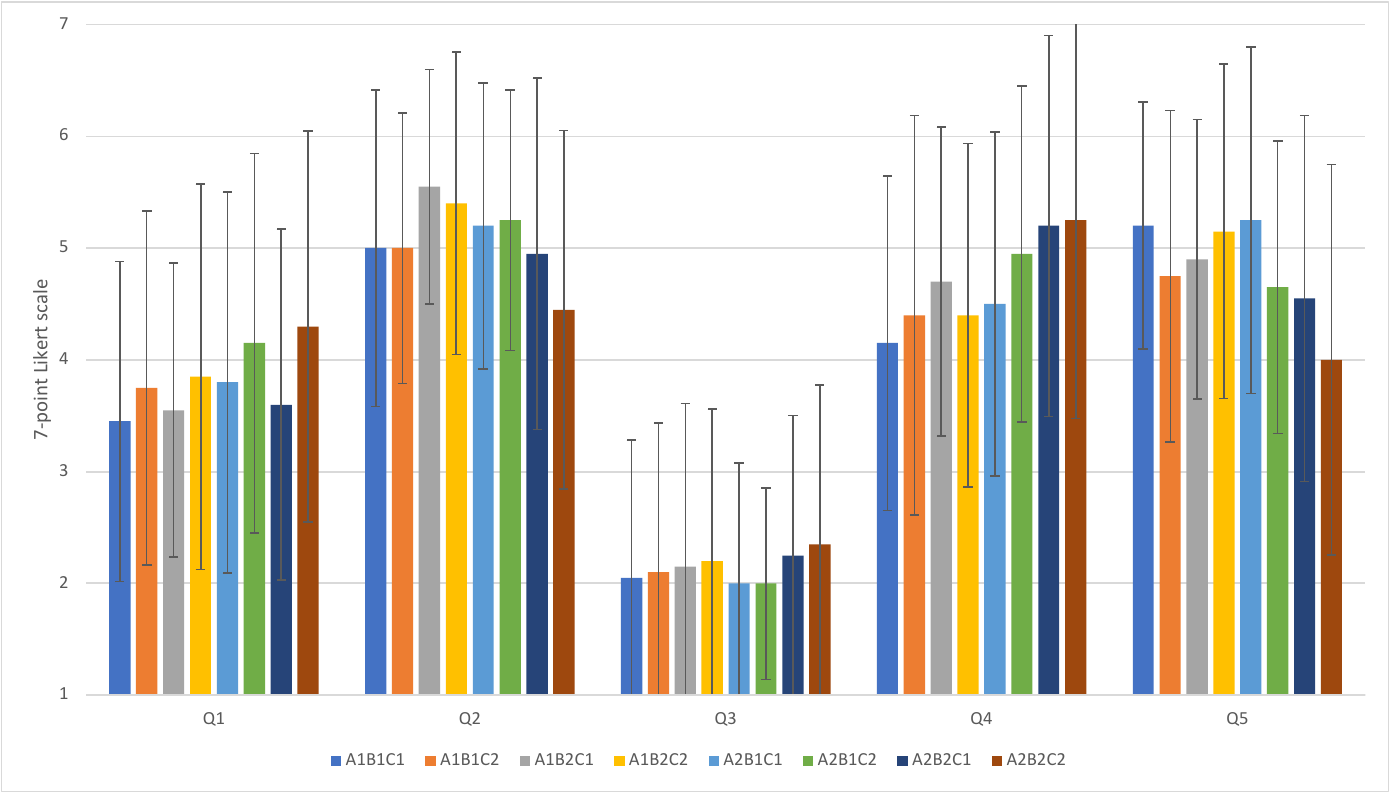}
    \caption{Mean and Standard Deviation (Experiment 2)}
    \label{test2_graph}
  \end{center}
\end{figure*}

\noindent\textbf{Results: }A three-way analysis of variance (ANOVA) was performed on the response data obtained from the experiment for each question item, using a significance level of $a$ = 0.05. The results of the analysis are shown in Table \ref{table:anova1}, Table \ref{table:anova2} and Figure \ref{test2_graph}.

\begin{table}[t]
  \centering
  \caption{Results of the ANOVA}
  \label{table:anova1}
  \footnotesize
  \setlength{\tabcolsep}{8pt}
  \begin{tabular}{l cccccc}
    \toprule
    & \multicolumn{2}{c}{Factor A} & \multicolumn{2}{c}{Factor B} & \multicolumn{2}{c}{Factor C} \\
    \cmidrule(lr){2-3} \cmidrule(lr){4-5} \cmidrule(lr){6-7}
    & $F$ & $p$ & $F$ & $p$ & $F$ & $p$ \\
    \midrule
    Q1 & 1.748 & .202  & 0.025 & .875  & 5.373 & .032* \\
    Q2 & 2.423 & .136  & 0.017 & .869  & 1.443 & .244  \\
    Q3 & 0.079 & .781  & 2.789 & .111  & 0.226 & .640  \\
    Q4 & 5.318 & .033* & 4.498 & .047* & 0.293 & .595  \\
    Q5 & 3.581 & .074  & 2.436 & .135  & 4.061 & .058  \\
    \bottomrule
    \multicolumn{7}{l}{\scriptsize *$p < .05$}
  \end{tabular}
\end{table}

\begin{table*}[t]
  \centering
  \caption{Results of the ANOVA (two-way interactions)}
  \label{table:anova2}
  \footnotesize
  \setlength{\tabcolsep}{6pt}
  \begin{tabular}{l ccc ccc ccc}
    \toprule
    & \multicolumn{3}{c}{Interaction AB}
    & \multicolumn{3}{c}{Interaction AC}
    & \multicolumn{3}{c}{Interaction BC} \\
    \cmidrule(lr){2-4}
    \cmidrule(lr){5-7}
    \cmidrule(lr){8-10}
    & $F$ & $p$ & Simple effect
    & $F$ & $p$ & Simple effect
    & $F$ & $p$ & Simple effect \\
    \midrule
    Q1
      & 0.172 & .683  & --
      & 0.207 & .654  & --
      & 0.117 & .736  & -- \\
    Q2
      & 12.459 & .002* & A(b2), B(a1)
      & 0.145  & .708  & --
      & 1.223  & .283  & -- \\
    Q3
      & 0.835 & .372  & --
      & 0.000 & 1.000 & --
      & 0.068 & .797  & -- \\
    Q4
      & 0.151 & .702  & --
      & 0.345 & .564  & --
      & 1.275 & .273  & B(c1) \\
    Q5
      & 3.209 & .089  & --
      & 1.498 & .236  & A(c2), C(a2)
      & 1.115 & .304  & B(c1), C(b1) \\
    \bottomrule
    \multicolumn{10}{l}{\scriptsize *$p < .05$}
  \end{tabular}
\end{table*}

\section{Discussion}
The results of this study indicate that the presence and patterns of finger movements exert a psychological influence on the user. It can be said that these finger movements functioned as non-verbal information complementing the robot's intent, playing a role in having the user perceive the robot not merely as a machine, but as an agent with its own intentions.

Repeated finger movements intensified the user's sense of being wanted; this is likely because the repetitive action of releasing and grasping the fingers served as an expression of affection. While asynchronous movements further heightened this sensation by creating the impression of an impulsive desire for contact, simultaneous movements were perceived as rational actions that evoked feelings of care and consideration.

The strength of the embrace also played a role. A strong embrace was conveyed as a childlike expression of affection—suggesting a lack of control over the force applied—whereas a weak embrace was interpreted as a considerate gesture aimed at minimizing physical strain.

\section{Conclusion}
In this study, we developed a hugging robot and conducted experiments to clarify the psychological effects of finger movements and the intensity of the embrace. The results suggest that while finger movements foster a sense of being wanted by the user, a strong embrace influences impressions of childlike qualities and thoughtfulness.

Beyond these findings, deeper emotional engagement through hugging raises ethical issues regarding the robot's intentions, moral treatment, and the locus of responsibility—such as the risk of excessive dependency when a robot comforts a child. Furthermore, delicate physical contact risks being perceived as overly intimate or sexually suggestive, potentially undermining mutual psychological safety.

Consequently, future designs may need to intentionally limit the depth of interaction. Equipping robots with self-protective behaviors—such as the ability to refuse inappropriate hugs—will likely be essential for maintaining safe and appropriate boundaries between humans and robots.

\begin{acks}
This research was supported in part by JSPS KAKENHI Grant Numbers 23K11278, 26K14911, 23K11202, and 24K02977.
\end{acks}

\bibliographystyle{ACM-Reference-Format}
\bibliography{reference}

\appendix

\end{document}